\documentclass[wcp]{jmlr}

\usepackage{longtable}

\usepackage{booktabs}

\usepackage{lineno}

\usepackage{longtable}
\usepackage{booktabs}
\usepackage{lineno}
\usepackage{microtype}
\usepackage{hyperref}
\usepackage{xcolor}
\usepackage{bm}
\usepackage{makecell}
\usepackage{algpseudocode}
\usepackage{bbm}
\usepackage{dsfont}
\usepackage{subcaption}
\usepackage{colortbl}

\definecolor{LightGreen}{RGB}{230,255,230}
\definecolor{LightRed}{RGB}{255,230,230}
\definecolor{claimBlue}{RGB}{184,230,247}  
\newcommand{\good}[1]{\cellcolor{LightGreen}{#1}}

\definecolor{darkblue}{rgb}{0, 0, 0.5}
\hypersetup{colorlinks=true, citecolor=darkblue, linkcolor=darkblue, urlcolor=darkblue}

\makeatletter
\let \@jmlrpages \@empty
\makeatother

\jmlrvolume{353,}
\jmlryear{2026}
\jmlrworkshop{ACML 2026}

\title[Adaptive Conformal for LLMs]{Adaptive Conformal Prediction for Improving Factuality \\ of Generations by Large Language Models}

 \author{\Name{Aleksandr Rubashevskii} \Email{Aleksandr.Rubashevskii@mbzuai.ac.ae}\\
 \Name{Dzianis Piatrashyn} \Email{Dzianis.Piatrashyn@mbzuai.ac.ae}\\
 \Name{Preslav Nakov} \Email{Preslav.Nakov@mbzuai.ac.ae}\\
 \Name{Maxim Panov} \Email{Maxim.Panov@mbzuai.ac.ae}\\
\addr Mohamed Bin Zayed University of Artificial Intelligence (MBZUAI), Abu Dhabi, UAE}

\editors{Andy Song, Bo Han and Sarah Erfani}

\begin{document}

\makeatletter
\let \@jmlrpages \@empty
\makeatother

\maketitle

\begin{abstract}
  Large language models (LLMs) are prone to generating factually incorrect outputs.
  Recent work has applied conformal prediction to provide uncertainty estimates and statistical guarantees for the factuality of LLM generations.
  However, existing approaches are typically not prompt-adaptive, limiting their ability to capture input-dependent variability.
  As a result, they may filter out too few items (leading to over-coverage) or too many (under-coverage) for a given task or prompt.
  We propose an adaptive conformal prediction approach that extends conformal score transformation methods to LLMs, with applications to long-form generation and multiple-choice question answering.
  This enables prompt-dependent calibration, retaining marginal coverage guarantees while improving conditional coverage.
  In addition, the approach naturally supports selective prediction, allowing unreliable claims or answer choices to be filtered out in downstream applications.
  We evaluate our approach on multiple white-box models across diverse domains and show that it significantly outperforms existing baselines in terms of conditional coverage.
  Our implementation is available at \url{https://github.com/stat-ml/adaptive_conformal_factuality}.
\end{abstract}


\section{Introduction}
\label{sec:intro}

Large language models (LLMs) have demonstrated impressive performance across diverse applications~\citep{zhao2023survey, minaee2024large}.
Despite this progress, they are still susceptible to hallucinations, producing fluent but factually incorrect outputs~\citep{huang2025survey}.
This limitation is especially concerning in high-risk domains such as medicine, where even a few errors within extended generations can lead to significant consequences~\citep{thirunavukarasu2023large}.

To mitigate these risks, it is essential to develop methods with rigorous reliability guarantees.
Conformal prediction offers a theoretically grounded approach to uncertainty quantification, providing distribution-free guarantees on error rates~\citep{vovk2005algorithmic, angelopoulos2023conformal}.

Conformal prediction has recently been applied to large language models in tasks such as long-form question answering~\citep{mohri2024language} and multi-choice QA~\citep{kumar2023conformal}. In these settings, conformal methods are typically used to construct prediction sets or filtering rules based on uncertainty scores, enabling selective prediction: the model either returns only high-confidence outputs or abstains from uncertain ones.
For example, in long-form generation, individual claims or spans can be filtered based on their estimated reliability, while in multiple-choice settings, conformal prediction produces a subset of candidate answers guaranteed to contain the correct one with high probability.

However, existing conformal procedures for LLMs lack adaptivity: a single calibrated quantile is applied uniformly across all test prompts, regardless of their difficulty, ambiguity, or rarity.
While this procedure guarantees marginal coverage on average, it can lead to substantial miscalibration at the prompt level~\citep{cherian2024large}.
For certain inputs, the method may exhibit over-coverage (an overly conservative threshold), whereas for others it may result in under-coverage (an insufficiently strict threshold).

We propose an adaptive conformal prediction approach for evaluating the factuality of large language models that accounts for the characteristics and difficulty of specific tasks. 
The proposed methodology is evaluated across multiple domains using various models and uncertainty quantification techniques.
Figure~\ref{fig:long} illustrates the main result of our work: standard conformal methods fail to achieve category-wise (conditional) coverage for heterogeneous prompts, whereas our adaptive approach improves conditional coverage while preserving marginal guarantees (see the details of the experiment in Section~\ref{sec:experiments}).

Our contributions are as follows:
\begin{enumerate}
    \item We propose a new conformal prediction approach for hallucination detection in LLMs that learns a prompt-adaptive correction to conformity scores via embedding-conditioned quantile regression.
    
    \item Our method preserves the finite-sample marginal coverage guarantees of split conformal prediction while improving conditional coverage across heterogeneous
    prompts.
    
    \item Experiments on long-form and multiple-choice question answering benchmarks across multiple LLMs show improved hallucination detection performance and more stable coverage compared to existing conformal methods.
\end{enumerate}


\section{Methodology}
\subsection{Background}

  \begin{figure*}[t!]
    \centering
    \begin{minipage}[t]{0.43\textwidth}
      \includegraphics[width=\textwidth]{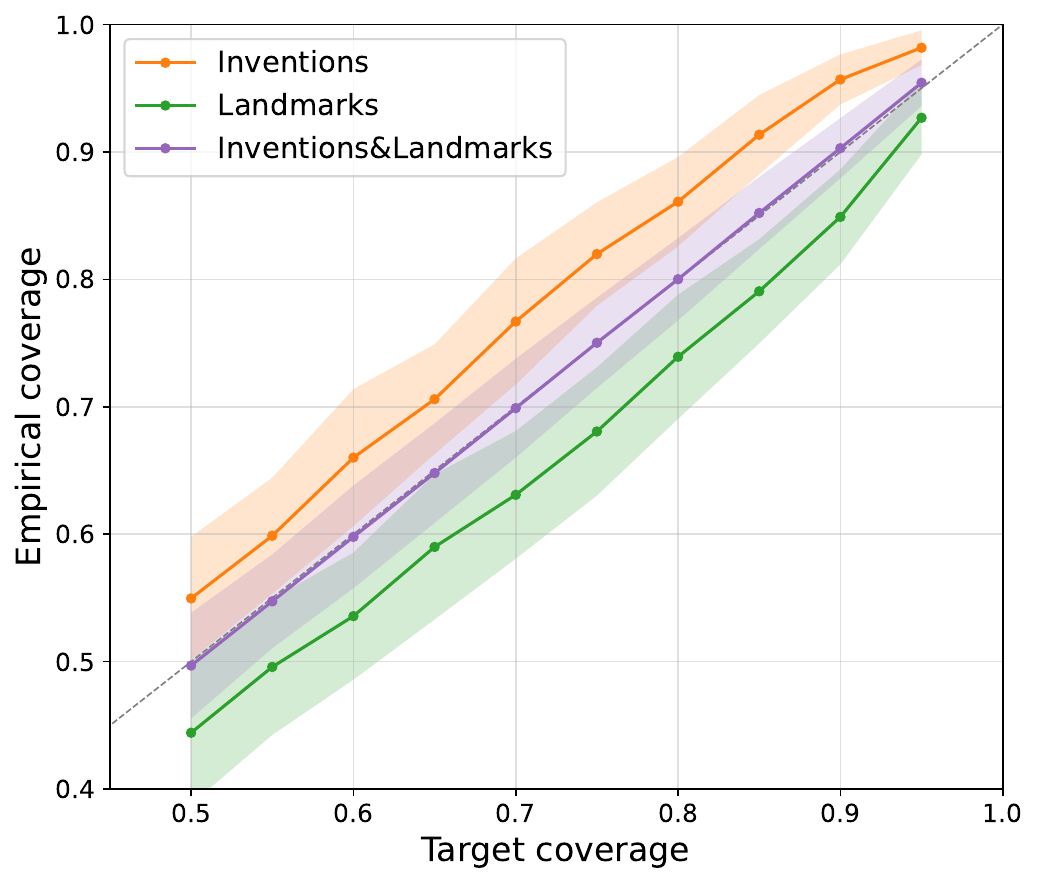}
      \vskip-7pt
      \centerline{(a)}
    \end{minipage}
    ~~~~
    \begin{minipage}[t]{0.43\textwidth}
      \includegraphics[width=\textwidth]{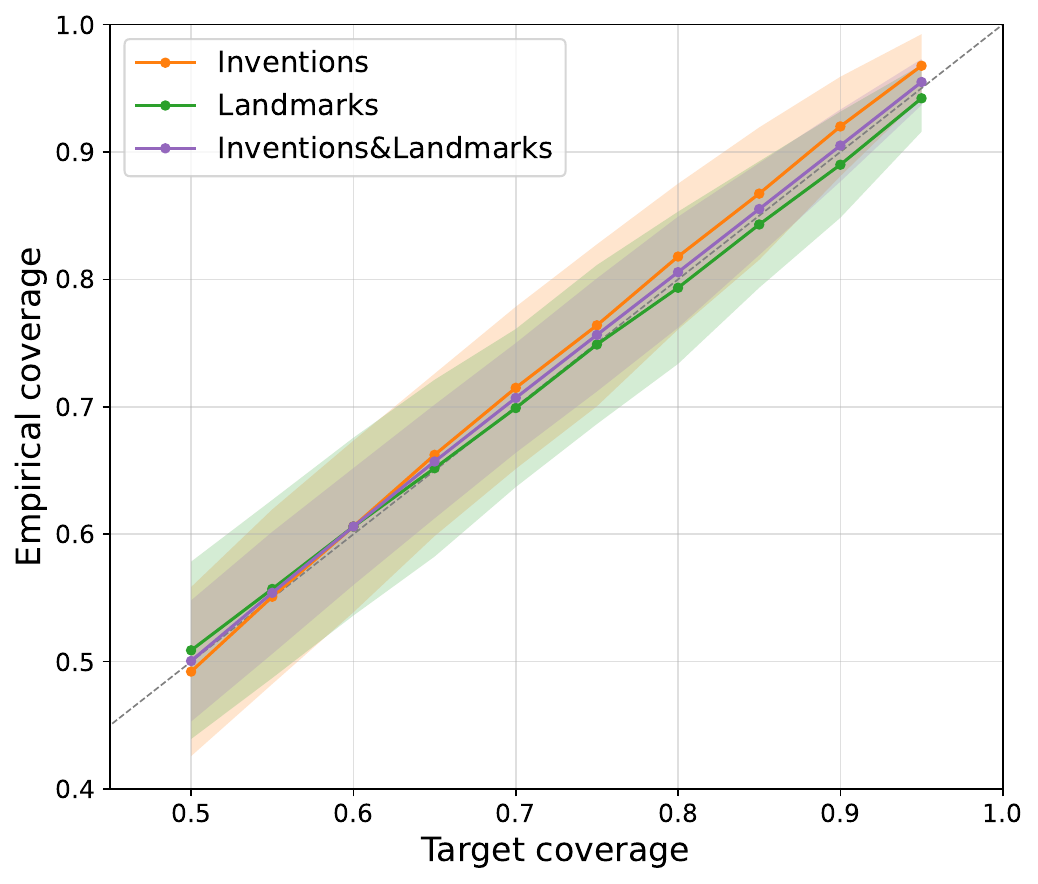}
      \vskip-7pt
      \centerline{(b)}
    \end{minipage}
    \vskip-5pt
    \caption{Long-form QA target vs. empirical coverage for (a) Conformal Factuality and (b) Adaptive Conformal (ours).
    The conformal procedure is calibrated jointly on two categories (``Inventions'' and ``Landmarks'').
    While both methods achieve the target coverage marginally, only the adaptive approach closely approximates conditional coverage.
    }
    \vskip-10pt
  \label{fig:long}
  \end{figure*}

  Conformal prediction
  assumes exchangeable data $\{ X_i, Y_i \}_{i=1}^{N+1}$ with input features $X_i$ and output labels $Y_i$, and a user-specified miscoverage level $\alpha$.
  Using calibration data $\{X_i, Y_i\}_{i=1}^{N}$ it constructs a prediction set $C_\alpha(x)$ such that for a new test point $\{X_{N+1}, Y_{N+1}\}$:
  \begin{equation}
    \mathbb{P} \bigl(Y_{N + 1} \in \mathcal{C}_\alpha (X_{N + 1})\bigr) \geq 1-\alpha .
    \label{eq:conformal_guarantees}
  \end{equation}
  This guarantee is marginal coverage, meaning the coverage holds on average over the distribution of $X_{N+1}$.

  Usually, conformal prediction approaches assume that some predictive model $\widehat{f}(x)$ was constructed that models the dependence between $x$ and $y$. Let $V(x, y)$ be a nonconformity score function, where larger values indicate worse agreement between $\widehat{f}(x)$ and $y$. 
  Using a calibration set $\{ X_i, Y_i \}_{i=1}^{N}$, define the calibration scores $V(X_i, Y_i), i = 1, \ldots, N$.
  Then the conformal prediction set is 
    \begin{equation}
    \mathcal{C}_\alpha(x) = \big\{ y \in \mathcal{Y}\colon V(x,y) \le Q_{1-\alpha}\big(\{V(X_i, Y_i)\}_{i=1}^N\big) \big\},
    \label{eq:pred_set}
    \end{equation}
    where $Q_{1-\alpha}\big(\{V(X_i, Y_i)\}_{i=1}^N\big)$ denotes the conformal $(1-\alpha)$-quantile: the $\lceil (N+1)(1-\alpha) \rceil$-th smallest element of $\{V(X_i, Y_i)\}_{i=1}^N$, or $+\infty$ if $\lceil (N+1)(1-\alpha) \rceil > N$.

\subsection{Conformal Prediction for LLMs}
\label{sec:conform_llm}
  Since LLMs generate free-form text, fine-grained factuality assessment typically decomposes responses into atomic, verifiable claims.
  For example, ``Paris is the capital of France and was founded in the 3rd century BC'' yields the claims ``Paris is the capital of France'' and ``Paris was founded in the 3rd century BC''.

  For input $x$, the LLM produces atomic, verifiable claims $L(x) = \{c_1, \ldots, c_m\} \subset \mathcal{C}$, where $\mathcal{C}$ is the space of claims.
  Since uncertain claims are more likely to be incorrect, given an uncertainty score $s\colon \mathcal{C} \to \mathbb{R}$ (higher means more uncertain), we define the filtered output at threshold $t$ as
  \begin{equation}
    F_t\bigl(x, L(x)\bigr) := \{c \in L(x)\colon s(c) \le t \}.
    \label{eq:filtration}
  \end{equation}
  Intuitively, $F_t$ retains only sufficiently low-uncertainty (i.e., confident) claims.
  Accordingly, the filtered set of claims produced by a large language model can be interpreted as a conformal prediction set, such as $C_\alpha$ in equation~\eqref{eq:pred_set}, as it restricts the output space to claims whose uncertainty scores do not exceed a calibrated threshold.

  Let $w\colon \mathcal{C} \times \mathcal{Y} \to \mathbb{R}$ be a claim-level factuality function (e.g., based on a pre-trained Natural Language Inference (NLI) model) that evaluates whether a claim is supported by the reference.
  We distinguish this from an uncertainty score $s(c)$, which provides a model-based estimate of how likely a claim is to be incorrect and is used to rank and filter claims.
  In contrast, $w(c,y)$ serves as an oracle that determines whether a claim is factually correct with respect to the ground-truth answer $y$.
  

  For the long-form QA setting, we define the score $V(x,y)$ as the largest uncertainty threshold such that all retained claims are factually correct:
  \begin{equation}
    V(x,y) = \sup \left\{t\colon \forall c \in F_t(x,L(x)),\; w(c,y) \ge \beta \right\},
    \label{eq:score_long}
  \end{equation}
  where $\beta$ is a fixed task-dependent factuality threshold defining claim correctness.




We compute the conformal threshold from the scores $\{ V(X_i,Y_i)\}_{i=1}^N$ on a
calibration set. Unlike the nonconformity score in~(2), $V$ is larger when more claims
can be retained safely: $F_t(x,L(x))$ is factually correct if and only if $t < V(x,y)$.
The $(1-\alpha)$-quantile is therefore taken from the opposite end: for the scores~(4),
$Q_{1-\alpha}(\{ V(X_i,Y_i) \}_{i=1}^N)$ is the $\lceil (N+1)(1-\alpha) \rceil$-th largest of
$\{ V(X_i,Y_i) \}_{i=1}^N$,
or $-\infty$ if this index exceeds $N$. At test time,
\begin{equation}
  \bar L_\alpha(x) = \big\{ c \in L(x)\colon s(c) \le Q_{1-\alpha}\big(\{ V(X_i,Y_i) \}_{i=1}^N\big) \big\}.
  \label{eq:long_set}
\end{equation}


  %
  This procedure ensures the following marginal coverage guarantee:
  \begin{equation}
    \mathbb{P} \Bigl(\forall c \in \bar{L}_{\alpha}(X_{N+1})\colon w(c, Y_{N+1}) \ge \beta \Bigr) \ge 1-\alpha.
  \label{eq:guaranties}
  \end{equation}
  The condition $w(c,Y)\ge \beta$ plays a role analogous to the membership test $Y \in C_{\alpha}(X)$ in classical conformal prediction, defining whether a retained prediction is correct; see equation~\eqref{eq:conformal_guarantees}.
  \citet{mohri2024language} propose a related mechanism for long-form question answering using entailment-based sets.

  \paragraph{Multiple-Choice QA Setting.} We further note that the reformulation of conformal prediction for long-form QA naturally extends to the multi-choice QA setting.
  In this case, the elements $c$ in equation~\eqref{eq:filtration} correspond to candidate answer classes, and the filtration mechanism $F_t$ produces a subset of predicted classes.
  The factuality function in equation~\eqref{eq:guaranties} reduces to verifying whether the true class $Y_{N+1}$ is contained in the filtered set $\bar{L}_{\alpha}(X_{N+1})$.
  In this setting, the nonconformity score $V(x,y)$ can be defined using the least ambiguous classifier (LAC; \citealp{sadinle2019least}):
  \begin{equation}
    V(x, y) = 1 - [p(x)]_{y} \; ,
    \label{eq:score_mc}
  \end{equation}
  where $[p(x)]_{y}$ denotes the predicted probability of the true class $y$.
  Under this formulation, the same marginal coverage guarantee is recovered: the true class belongs to the constructed prediction set with probability at least $1 - \alpha$.

\subsection{Adaptive Conformal Prediction}
  Standard conformal prediction methods for LLMs rely on global thresholds and do not account for input-dependent variability, which can lead to substantial over- or under-coverage for specific inputs despite valid marginal guarantees.
  To address this limitation, we build on a class of methods that improve conditional coverage by transforming nonconformity scores using input-dependent normalization (see Section~\ref{sec:rel_work} for an overview of related works).
  
  In this framework, the transformed score is defined as
  \begin{equation}
    \tilde{V}(x,y) = f^{-1}_{\tau(x)}\bigl(V(x,y)\bigr),
  \end{equation}
  where $\tau(x)$ is an estimate of the conditional $(1-\alpha)$-quantile of the original score, taken from the same end as $Q_{1-\alpha}$.
  Such transformations aim to normalize the score so that its conditional quantiles are approximately invariant with respect to $x$, aligning the distributions across inputs.
  In this work, we consider a simple multiplicative normalization given by division by the estimated conditional quantile.
  This corresponds to the choice $f_t(v) = t \cdot v$, for which $f_t^{-1}(v) = v/t$. 
  This transformation can be interpreted as a local rescaling, reducing variability of the score across inputs and bringing conditional quantiles closer together.

  More generally, other transformations are possible within this framework.
  For example, additive normalization via shifting the score by its estimated conditional quantile can similarly reduce input dependence of the relevant quantile.

  Conformal prediction sets are then constructed using the transformed scores:
  \begin{equation}
    \tilde{\mathcal{C}}_\alpha(x) = \left\{ y \in \mathcal{Y}\colon \tilde{V}(x, y) \le 
    Q_{1-\alpha}\big(\{\tilde V(X_i, Y_i) \}_{i=1}^N\big) \right\}.
  \end{equation}

  This class of score-transformation methods has primarily been studied in regression settings and evaluated on relatively small-scale datasets.
  In contrast, we extend this framework to long-form LLM generation, where outputs consist of multiple sentences and atomic claims.
  In this setting, achieving approximate conditional validity is more challenging due to the need for large calibration data, informative input representations (e.g., prompt embeddings), and the complexity of long-form outputs.

\subsection{Adaptive Conformal Factuality}
\paragraph{Long-form QA.}
  Dataset $\mathcal{D}$ consists of prompt–generation pairs $\big(x, L(x)\big)$, where the model output 
  $L(x) = \{c_1, \ldots, c_m\}$ is a set of extracted verifiable atomic claims. 
  For each prompt $X_i$, $i=1,\ldots,n$, we additionally compute a sentence embedding $e(X_i)$.
  We split the dataset $\mathcal{D}$ into three disjoint subsets: $\mathcal{D}_{\mathrm{cal}_1}$, $\mathcal{D}_{\mathrm{cal}_2}$, and $\mathcal{D}_{\mathrm{test}}$.

  We build on the filtration mechanism $F_t$ and factuality function $w$ introduced in Section~\ref{sec:conform_llm}. 
  For the long-form QA setting, we define $V(x,y)$ as the maximal uncertainty threshold such that all retained claims are factually correct; see equation~\eqref{eq:score_long}.
  In our setting, factuality is evaluated using binary labels, so $w(c, y) \in \{0,1\}$ and $\beta = 1$, meaning that all retained claims must be correct.

  \begin{algorithm}[!ht]
    
    \hrule height 0.8pt\kern2pt

    \caption{Adaptive Conformal Factuality for Long-Form QA}
    \label{alg:long_form}
    
    \kern2pt\hrule height 0.8pt
    
    \KwIn{LLM $L$, miscoverage level $\alpha$, calibration sets $\bigl\{\bigl(X_i^{(1)},Y_i^{(1)}\bigr)\bigr\}_{i=1}^{n_{\mathrm{cal}_1}}$ and $\bigl\{\bigl(X_i^{(2)},Y_i^{(2)}\bigr)\bigr\}_{i=1}^{n_{\mathrm{cal}_2}}$, pre-trained prompt embedding extractor $e$, 
    function $V(x, y$) from equation~\eqref{eq:score_long},
    claim-level score function $s(c)$, 
    test prompt $x$.
    }
    \For{$i \gets 1$ \KwTo $n_{\mathrm{cal}_1}$}{
      $z^{(1)}_i \gets e\bigl(X_i^{(1)}\bigr)$\;
      
      $V_i^{(1)} \gets V\bigl(X_i^{(1)}, Y_i^{(1)}\bigr)$\;
    }

    Fit a conditional $(1-\alpha)$-quantile regressor $\hat{\tau}$ on $\bigl\{\bigl(z^{(1)}_i, V_i^{(1)}\bigr)\bigr\}_{i=1}^{n_{\mathrm{cal}_1}}$\;

    \For{$i \gets 1$ \KwTo $n_{\mathrm{cal}_2}$}{
      $z^{(2)}_i \gets e\bigl(X_i^{(2)}\bigr)$\;
      
      $\hat{\tau}^{(2)}_i \gets \hat{\tau}\bigl(z^{(2)}_i\bigr)$\;
      
      $\tilde V_i^{(2)} \gets \frac{V\bigl(X_i^{(2)}, Y_i^{(2)}\bigr)}
      {\hat{\tau}^{(2)}_i}
      $\;
    }
    $\hat q_{1-\alpha} \gets Q_{1-\alpha} \left( \bigl\{
    \tilde V^{(2)}_i
    \bigr\}_{i=1}^{n_{\text{cal}_2}} \right)$\;

    \KwOut{$\bar{L}_{\alpha}(x) \gets \{ c \in L(x)\colon \frac{s(c)}{\hat{\tau}(x)} \le \hat q_{1-\alpha} \}$}
    \hrule
  \end{algorithm}
  
  On $\mathcal{D}_{\mathrm{cal}_1}$, we compute scores $\bigl\{V\bigl(X^{(1)}_i, Y^{(1)}_i\bigr)\bigr\}_{i=1}^{n_{\mathrm{cal}_1}}$ and train a conditional quantile estimator $\hat{\tau}(x)$ (using the pinball loss) on the pairs
  $\bigl\{\bigl(e\bigl(X^{(1)}_i\bigr), V\bigl(X^{(1)}_i,Y^{(1)}_i\bigr)\bigr)\bigr\}_{i=1}^{n_{\mathrm{cal}_1}}$.
  We use $\hat{\tau}(x)$ as shorthand for $\hat{\tau}(e(x))$, where the conditional quantile is evaluated on the embedding $e(x)$.
  The details of this procedure are provided in Section~\ref{subsec:exp_setup}.

  On $\mathcal{D}_{\mathrm{cal}_2}$, we compute transformed scores $\bigl\{\tilde{V}\bigl(X^{(2)}_i,Y^{(2)}_i\bigr)\bigr\}_{i=1}^{n_{\mathrm{cal}_2}}$:
  \begin{equation}
    \tilde{V}\bigl(X^{(2)}_i,Y^{(2)}_i\bigr) = \frac{V\bigl(X^{(2)}_i,Y^{(2)}_i\bigr)}{\hat{\tau}\bigl(X^{(2)}_i\bigr)}.
  \label{eq:transform}
  \end{equation}
  We then compute the conformal threshold
  $Q_{1-\alpha}\bigl(
  \bigl\{\tilde{V}\bigl(X^{(2)}_i,Y^{(2)}_i\bigr)\bigr\}_{i=1}^{n_{\mathrm{cal}_2}}
  \bigr)$.
  Since $\hat\tau(x) > 0$, the transformed scores~\eqref{eq:transform} have the same orientation as $V$, so $Q_{1-\alpha}$ in~\eqref{eq:long_transform} 
  is taken from the same end as in~\eqref{eq:long_set}.

  At test time, we evaluate transformed scores of candidate claims and filter them using the calibrated threshold. 
  Both the claim-level scores $s(c)$ and the calibration thresholds $V(x,y)$ are normalized by $\hat{\tau}(x)$, so that they are expressed on the same scale and can be compared using a single global threshold.
  The resulting conformal prediction set is
  \begin{equation}
    \bar{L}_{\alpha}(x)
    =
    \left\{
    c \in L(x)\colon
    \dfrac{s(c)}{\hat{\tau}(x)}
    \le
    Q_{1-\alpha} \left(\{\tilde{V}(X^{(2)}_i,Y^{(2)}_i)\}_{i=1}^{n_{\mathrm{cal}_2}} \right)
    \right\}.
    \label{eq:long_transform}
  \end{equation}

  The resulting algorithm is summarized in Algorithm~\ref{alg:long_form}.
  The predictions satisfy marginal coverage guarantees as in equation~\eqref{eq:guaranties}.
  The corresponding theoretical result and its proof are provided in Appendix~\ref{app:theorem}.

\paragraph{Multi-choice QA.}
  The proposed method also applies to multiple-choice question answering. 
  The same pipeline is used: training the conditional quantile estimator on $\mathcal{D}_{\mathrm{cal}_1}$, calibrating transformed scores on $\mathcal{D}_{\mathrm{cal}_2}$, and filtering on $\mathcal{D}_{\mathrm{test}}$.
  The main differences are:
  (i) the prediction set $\bar{L}_{\alpha}(x)$ consists of classes rather than claims,
  (ii) the task-specific nonconformity score is given by the least ambiguous classifier (see equation~\eqref{eq:score_mc}).
  The resulting conformal prediction set is
  \begin{equation}
    \bar{L}_{\alpha}(x) = \left\{y \in \mathcal{Y}\colon \frac{V(x, y)}{\hat{\tau}(x)} \le 
    Q_{1-\alpha} \left(\bigl\{\tilde{V}\bigl(X^{(2)}_i,Y^{(2)}_i\bigr)\bigr\}_{i=1}^{n_{\mathrm{cal}_2}} \right)
    \right\}.
  \end{equation}


\section{Experimental Study}
\label{sec:experiments}

\subsection{Setup}
\label{subsec:exp_setup}
\begin{figure*}[t!]
  \centering
  \begin{minipage}[t]{0.43\textwidth}
    \includegraphics[width=\textwidth]{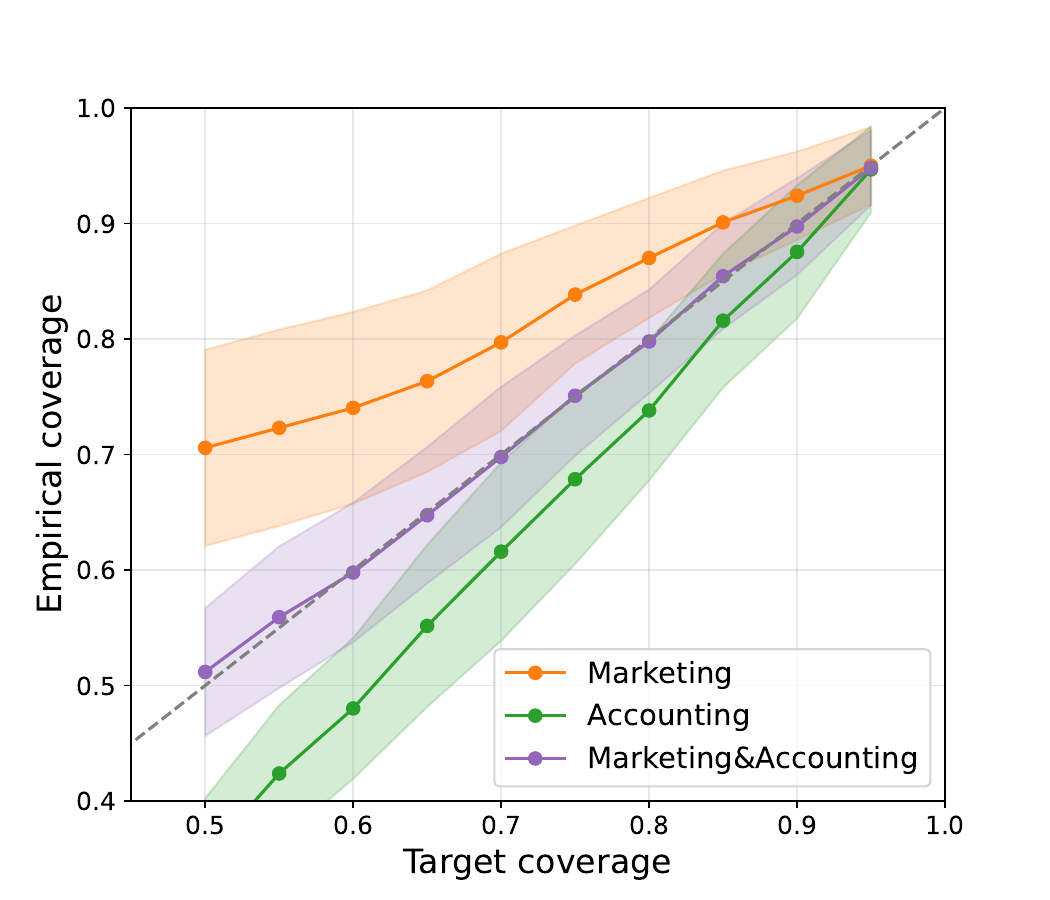}
    \vskip-10pt
    \centerline{(a)}
  \end{minipage}
  ~~~~
  \begin{minipage}[t]{0.43\textwidth}
    \includegraphics[width=\textwidth]{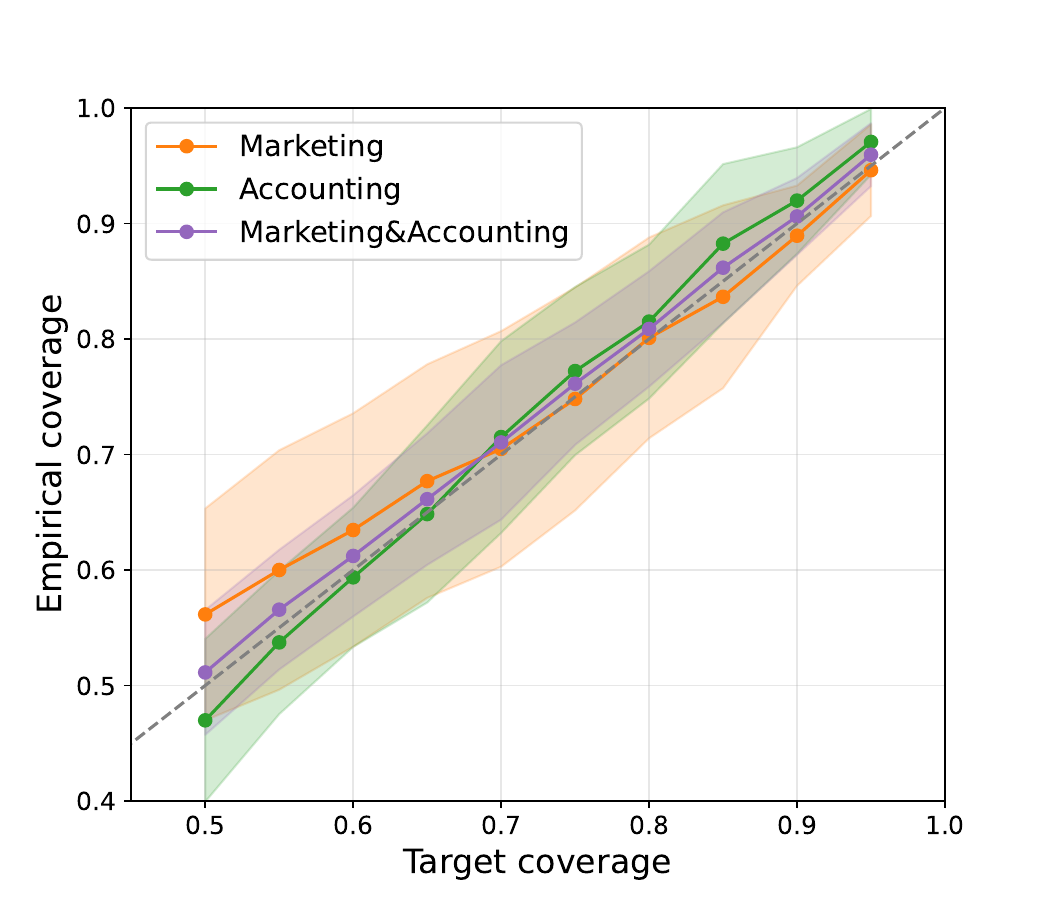}
    \vskip-10pt
    \centerline{(b)}
  \end{minipage}
  \vskip-5pt
  \caption{Multi-choice QA experimental results: target vs. empirical coverage for (a) Conformal Factuality and (b) Adaptive Conformal method (ours).
  Results are shown for two prompt categories, with coverage reported for each category individually as well as jointly. 
  The conformal threshold is calibrated jointly across both categories.
  }
  \vskip-10pt
\label{fig:mcqa_two_classes}
\end{figure*}

In the main experiments, generations are produced by Mistral-7B-Instruct-v0.2~\citep{jiang2023mistral7b}, Llama-3.1-8B-Instruct~\citep{grattafiori2024llama} and Gemma-3-12B-Instruct~\citep{gemmateam2025gemma3technicalreport}.
The illustrative MCQA experiments use Llama-2-7B~\citep{touvron2023llama}.

We extract prompt embeddings using the multi-qa-mpnet-base-dot-v1 model~\citep{reimers-2020-multilingual-sentence-bert} and reduce the resulting $768$-dimensional embeddings to $32$ dimensions via PCA.
We split the data $\mathcal{D}$ into three disjoint subsets $\mathcal{D}_{\mathrm{cal}_1}$, $\mathcal{D}_{\mathrm{cal}_2}$ and $\mathcal{D}_{\mathrm{test}}$ in proportions $0.3$, $0.4$ and $0.3$, respectively.
The conditional quantile is modeled by a small MLP with ReLU activation, trained with the pinball loss on $\mathcal{D}_{\mathrm{cal}_1}$.
We repeat each experiment 100 times for long-form QA and 20 times for multiple-choice QA.
In each run, we randomly shuffle the data and create a new split.
We report the mean and standard deviation over all runs.
Further details on the experimental design are provided in Appendix~\ref{subsec:ablation}.


\subsection{Dataset}
\subsubsection{Long-form QA}
Following \cite{shelmanov2025head}, we generate long-form samples for each of the $8$ categories: Biographies, Cities, Movies, Inventions, Books, Artworks, Landmarks, and Events.
All generations are decomposed and decontextualized into atomic claims, which are subsequently labeled using GPT-4o.
Instead of generating the original $100$ samples per category, we produce three times as many, resulting in $300$ long-form LLM generations per category.

The motivation for increasing the sample size is that we aim to provide per-prompt conformal guarantees on factuality.
Moreover, the data is further divided into three disjoint subsets for conditional quantile training, calibration, and testing.
Consequently, several hundred samples per category are required to obtain representative and robust estimates.


As for the claim scoring function $s(c)$, we consider several claim-level uncertainty measures for white-box models, including Maximum Probability, Maximum Token Entropy~\citep{fomicheva-etal-2020-unsupervised}, Perplexity~\citep{fomicheva-etal-2020-unsupervised}, Claim Condition Probability~\citep{fadeeva2024fact},
TokenSAR~\citep{duan2024shifting}, Pointwise Mutual Information~\citep{takayama-arase-2019-relevant}.
For data generation and claim-level uncertainty quantification, we use the LM-Polygraph library~\citep{fadeeva-etal-2023-lm,vashurin-etal-2025-benchmarking}.

\subsubsection{Multi-choice QA}
Similar to~\cite{kumar2023conformal} for multiple-choice question answering, we select $16$ categories from the MMLU dataset~\citep{hendrycks2020measuring}.
Each data category has at least $100$ questions, each question has $4$ possible answers.
Unlike the original paper, which applies conformal prediction independently within each category, we construct a single conformal predictor using data from all categories jointly and subsequently evaluate its performance separately for each category.

\subsection{Long-form QA Experimental Results}

\begin{table}[t]
\centering
\small
\begin{tabular}{lccc}
\toprule
Claim Scoring Method & Mistral 7B & Llama3 8B & Gemma3 12B \\
\midrule
Random Baseline & 0.189 & 0.166 & 0.138 \\
Maximum Probability & 0.273 & 0.281 & 0.180 \\
Perplexity & 0.255 & 0.257 & 0.162 \\
Max Token Entropy & 0.313 & 0.324 & 0.189 \\
Pointwise Mutual Information & 0.189 & 0.158 & 0.136 \\
Claim Conditioned Probability & \good{0.360} & \good{0.367} & \good{0.238} \\
TokenSAR & 0.288 & 0.286 & 0.182 \\
\bottomrule
\end{tabular}
\caption{PR-AUC for long-form QA claim scoring functions on Mistral 7B, Llama3 8B and Gemma3 12B. 
Higher is better, best method per column is colored.}
\label{table:pr_auc_unc}
\end{table}

\paragraph{Claim Scoring Functions Comparison.} 
First, we compare various claim-level uncertainty quantification methods for claim filtering.
We evaluate performance using PR-AUC, which is more informative in imbalanced settings and directly captures the precision–recall trade-off when filtering incorrect claims.

Table~\ref{table:pr_auc_unc} shows that the Claim Conditioned Probability (CCP) method achieves the best performance across all evaluated generation models.
By focusing on claim-specific uncertainty rather than non-task-relevant factors such as claim order or surface form variability, CCP consistently outperforms competing approaches.
Based on these results, we use CCP as the claim scoring method in subsequent conformal prediction experiments for long-form QA.

\paragraph{Calibration on Two Categories.}
We compare global quantile thresholding via Conformal Factuality~\citep{mohri2024language} with our adaptive conformal approach based on transformed scores.
Conformal Factuality applies a single quantile threshold computed jointly on $\mathcal{D}_{\mathrm{cal}_1}$ and $\mathcal{D}_{\mathrm{cal}_2}$, which is then used at test time.
In contrast, our method uses $\mathcal{D}_{\mathrm{cal}_1}$ to train a conditional quantile estimator and $\mathcal{D}_{\mathrm{cal}_2}$ to calibrate the transformed scores.

In a long-form QA experiment with Mistral~7B, we select two categories with substantially different conformity score distributions: Inventions and Landmarks.
As shown in Figure~\ref{fig:long}, both methods satisfy the marginal conformal guarantee.
However, global thresholding fails to achieve conditional coverage, resulting in over-coverage for Inventions and under-coverage for Landmarks.
In contrast, our adaptive conformal procedure preserves marginal coverage while achieving improved conditional coverage, yielding more consistent performance across categories as well as on the overall dataset.

\paragraph{Calibration Using All Data.}

For this experiment, we calibrate the threshold jointly across all eight categories.
In addition to the global threshold of Conformal Factuality (Original), we consider two baselines that require the category of the test prompt.
Group-wise performs group-conditional calibration, computing a separate threshold for each category, and serves as an oracle reference.
Boosted is the conditional conformal method of \citet{cherian2024large}, which boosts the claim scores and calibrates the threshold over predefined groups, here given by the category indicators.


Without access to category labels, the adaptive method substantially improves category-wise coverage, reducing the coverage bias (Cov.\ bias), i.e., the absolute deviation of the category-wise coverage from the target averaged over categories.
It is closer to the target than Boosted, which relies on category information, and recovers most of the gain of the oracle Group-wise calibration.
\citet{cherian2024large} enforce coverage over every predefined group, which can make the thresholds conservative in finite samples, consistent with the over-coverage of Boosted in our experiments.
Our method instead learns a prompt-dependent normalization from embeddings, without predefined groups, so that a single conformal quantile yields prompt-adaptive thresholds.
The fraction of removed correct claims remains comparable to that of the original method (Appendix~\ref{subsec:long_res}), at the cost of a higher variance across random splits.

\begin{table}[t]
\centering
\small
\setlength{\tabcolsep}{5pt}
\begin{tabular}{lcccc}
\toprule
& \multicolumn{2}{c}{Without category labels} & \multicolumn{2}{c}{With category labels} \\
\cmidrule(lr){2-3} \cmidrule(lr){4-5}
Category & Original & Adaptive & Group-wise & Boosted \\
\midrule
Inventions & 85.60 $\pm$ 3.34 & \good{81.03 $\pm$ 6.23} & 80.10 $\pm$ 5.22 & 82.41 $\pm$ 4.72 \\
Biographies & 81.39 $\pm$ 3.74 & \good{79.53 $\pm$ 5.51} & 79.67 $\pm$ 5.55 & 80.47 $\pm$ 5.65 \\
Artworks & 75.46 $\pm$ 4.49 & \good{81.06 $\pm$ 5.77} & 80.49 $\pm$ 5.16 & 80.69 $\pm$ 6.25 \\
Books & 84.34 $\pm$ 3.70 & \good{80.88 $\pm$ 5.60} & 80.43 $\pm$ 5.29 & 81.98 $\pm$ 5.61 \\
Cities & 79.25 $\pm$ 3.87 & \good{79.87 $\pm$ 6.16} & 80.02 $\pm$ 4.90 & 81.90 $\pm$ 4.78 \\
Movies & 83.57 $\pm$ 3.27 & \good{79.14 $\pm$ 6.28} & 80.82 $\pm$ 4.68 & 81.38 $\pm$ 6.10 \\
Landmarks & 73.34 $\pm$ 3.60 & \good{79.04 $\pm$ 5.34} & 80.09 $\pm$ 4.75 & 82.33 $\pm$ 5.85 \\
Events & 78.04 $\pm$ 3.28 & \good{81.54 $\pm$ 4.77} & 80.39 $\pm$ 4.48 & 81.87 $\pm$ 5.09 \\
\midrule
Cov. bias & 3.60 & \good{0.87} & 0.33 & 1.63 \\
\bottomrule
\end{tabular}
\caption{Mistral 7B results (mean $\pm$ std over random splits) for long-form QA with target coverage $80\%$.
Adaptive never sees the category of the test prompt, yet it comes closer to the target than both Original and Boosted, which does use categories, and reduces most of the gap to the oracle Group-wise calibration.
Color marks the label-free method closer to the target.}
\label{tab:mistral_ccp_merged}
\end{table}



\subsection{MCQA Experimental Results}

\paragraph{Calibration on Two Categories.}
We conduct an initial experiment on multiple-choice question answering using a setup analogous to the long-form QA setting.
Specifically, we select two categories out of the 16 available, namely Marketing and Accounting, which have substantially different nonconformity score distributions.
This experiment uses 40 random splits in proportions 0.4, 0.4 and 0.2.

Figure~\ref{fig:mcqa_two_classes} shows that while both methods achieve the desired marginal coverage overall, global conformal thresholding fails to provide accurate category-wise calibration.
In contrast, the adaptive conformal approach achieves coverage closer to the target for each category individually, demonstrating improved conditional coverage.
The relatively large variance reflects the inherent stochasticity of LLM outputs and their sensitivity to prompts; nevertheless, the adaptive method exhibits more stable behavior.

\begin{figure*}[t!]
  \centering
  \begin{minipage}[t]{0.45\textwidth}
    \includegraphics[width=\textwidth]{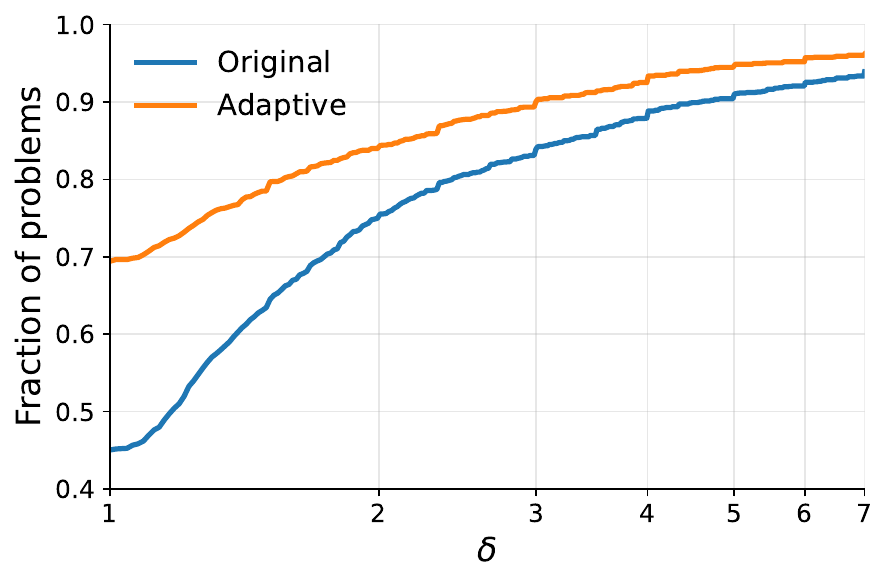}
    \vskip-10pt
    \centerline{(a)}
  \end{minipage}
  ~~~~
  \begin{minipage}[t]{0.45\textwidth}
    \includegraphics[width=\textwidth]{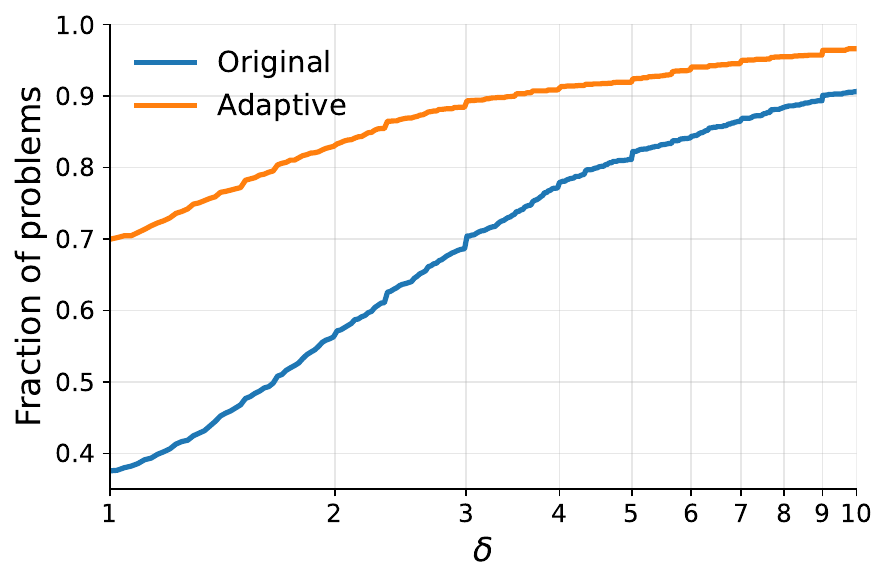}
    \vskip-10pt
    \centerline{(b)}
  \end{minipage}
  \vskip-5pt
  \caption{Dolan-Moré profiles for calibration error on MCQA for (a) Mistral 7B, (b) Llama3 8B.
  Problems are defined by (category, seed, $\alpha$) with $1 - \alpha \in \{0.5, 0.55, \dots, 0.8\}$, 20 seeds and 16 categories.
  Calibration error is defined as $|\text{empirical coverage} - (1-\alpha)|$, normalized per problem. 
  The x-axis ($\delta$) is plotted on a logarithmic scale. 
  Curves show the fraction of problems within a factor $\delta$ of the best (higher is better).}
  \vskip-10pt
\label{fig:dolan_more}
\end{figure*}

\paragraph{Calibration Using All Data.}
To compare calibration performance across all $16$ data categories, we use Dolan–Moré performance profiles~\citep{dolan2002benchmarking} for both the original and adaptive conformal methods.
Each problem instance is defined by a tuple (category, random seed, $\alpha$), where $1 - \alpha \in [0.5, 0.8]$ with step $0.05$, across $20$ seeds and $16$ categories.

We evaluate each method by its absolute deviation from nominal coverage and normalize performance relative to the best method on each problem.
Following the standard definition of Dolan–Moré profiles, we define the performance ratio
\noindent{\begin{equation}
    r_{ps} = \dfrac{t_{ps}}{\min_{s'} t_{ps'}},
\end{equation}}
where $t_{ps}$ denotes the coverage error of method $s$ on problem $p$.
This ratio measures how much worse a method performs compared to the best-performing method on a given problem.
Given a set of problems $\mathcal{P}$, the performance profile is defined as
\noindent{\begin{equation}
    \rho_s(\delta) = \frac{\left| \left\{ p \in \mathcal{P} \colon r_{ps} \le \delta \right\} \right|}{|\mathcal{P}|},
\end{equation}}
representing the fraction of problems for which method $s$ is within a factor $\delta$ of the best one.

Figure~\ref{fig:dolan_more} shows the resulting performance profiles.
Across both models, the adaptive method consistently outperforms the original method, with particularly strong gains for Llama3 8B, as evidenced by its uniformly higher curve across all values of $\delta$.
In particular, at $\delta=1$, it achieves the best calibration error on a larger fraction of problems, and remains closer to the best-performing method as $\delta$ increases.
Overall, this demonstrates more robust and reliable calibration across heterogeneous categories.

\begin{table}[t]
\centering
\small
\setlength{\tabcolsep}{5pt}
\begin{tabular}{lcccc}
\toprule
& \multicolumn{2}{c}{Without category labels} & \multicolumn{2}{c}{With category labels} \\
\cmidrule(lr){2-3} \cmidrule(lr){4-5}
Category & Original & Adaptive & Group-wise & Boosted \\
\midrule
Inventions & 86.07 $\pm$ 3.22 & \good{80.07 $\pm$ 6.14} & 79.32 $\pm$ 5.06 & 82.22 $\pm$ 4.83 \\
Biographies & 75.39 $\pm$ 4.07 & \good{79.76 $\pm$ 6.09} & 79.72 $\pm$ 4.73 & 82.54 $\pm$ 4.81 \\
Artworks & 65.11 $\pm$ 4.32 & \good{81.15 $\pm$ 5.81} & 80.00 $\pm$ 4.90 & 81.82 $\pm$ 6.15 \\
Books & 86.05 $\pm$ 3.06 & \good{80.98 $\pm$ 5.36} & 80.76 $\pm$ 4.42 & 82.11 $\pm$ 5.68 \\
Cities & 86.14 $\pm$ 3.31 & \good{81.44 $\pm$ 5.24} & 80.99 $\pm$ 4.95 & 82.15 $\pm$ 5.43 \\
Movies & 87.69 $\pm$ 3.45 & \good{80.03 $\pm$ 5.24} & 80.15 $\pm$ 4.66 & 81.17 $\pm$ 5.11 \\
Landmarks & \good{79.82 $\pm$ 3.50} & 79.33 $\pm$ 5.90 & 80.54 $\pm$ 4.85 & 82.17 $\pm$ 5.22 \\
Events & 74.04 $\pm$ 4.62 & \good{77.73 $\pm$ 6.49} & 79.42 $\pm$ 5.43 & 82.78 $\pm$ 5.35 \\
\midrule
Cov. bias & 6.45 & \good{0.85} & 0.50 & 2.12 \\
\bottomrule
\end{tabular}
\caption{Gemma-3 12B results (mean $\pm$ std over random splits) for long-form QA with target coverage $80\%$.
Color marks the label-free method closer to the target.}
\label{tab:gemma_ccp_merged}
\end{table}

\section{Related Work}
\label{sec:rel_work}
Recently, conformal prediction has been extended to large language models across several settings, including long-form generation~\citep{mohri2024language}, multiple-choice QA~\citep{kumar2023conformal}, and response sampling~\citep{quachconformal}.

More broadly, improving conditional coverage has been studied via input-dependent normalization of conformity scores. 
\cite{plassier2025rectifying} propose transforming scores to equalize conditional quantiles, closely related to normalized conformal prediction~\citep{johansson2021investigating, lei2018distribution}. 
Other approaches use localization or reweighting, including kernel-based methods~\citep{guan2023localized}, quantile regression forests~\citep{amoukou2023adaptive}, and learned score transformations~\citep{xie2024boosted}.

In the context of LLMs, \cite{cherian2024large} address conditional coverage via a boosting-based method that improves group-wise calibration. 
However, their approach relies on predefined groups and hand-crafted features, and requires solving a linear system for quantile estimation. 
In contrast, we achieve prompt-level adaptivity using learned representations and conditional quantile regression, without explicit grouping or feature engineering. 
A related direction considers domain-shift-aware conformal prediction, reweighting calibration samples based on similarity to test inputs~\citep{lin2025domain}.


\section{Conclusion}
  We propose a new adaptive conformal prediction framework for large language models based on nonconformity score transformations via conditional quantile regression.
  The method preserves marginal guarantees while enabling prompt-dependent calibration and improving conditional coverage.
  Experiments across multiple models and domains show consistent gains over existing baselines, particularly for heterogeneous categories.
  Future work includes extending adaptive conformal methods to broader generation tasks, improving input representations, and strengthening theoretical guarantees.


\bibliography{short_factuality_cr}

\appendix

\section{Theoretical Result}
\label{app:theorem}
 
Let $\hat\tau$ be a function fitted on $\mathcal{D}_{\mathrm{cal1}}$ only, so that it does not depend on $\{(X_k, Y_k)\}_{k=1}^{N+1}$, i.e., the $N$ calibration points from $\mathcal{D}_{\mathrm{cal2}}$ and the test point. 
The guarantee below holds for any $\hat\tau$ fitted on $\mathcal{D}_{\mathrm{cal1}}$ and its accuracy affects only conditional coverage.
For every $x$, let $f_{\hat\tau(x)}$ be a strictly increasing and continuous transformation
(for $f_t(v) = t\,v$ this requires $\hat\tau(x) > 0$), and define
\[
  V_k := V(X_k, Y_k), \qquad
  \tilde V_k := f^{-1}_{\hat\tau(X_k)}\big(V(X_k, Y_k)\big), \qquad k = 1, \dots, N+1.
\]
Define the prediction set
\[
  \bar L_\alpha(x) = \Big\{ c \in L(x)\colon f^{-1}_{\hat\tau(x)}\big(s(c)\big) \le \hat q \Big\},
\]
where $\hat q$ is the $\lceil (N+1)(1-\alpha) \rceil$-th largest of the calibration scores
$\tilde V_1, \dots, \tilde V_N$, i.e., $\hat q = Q_{1-\alpha}(\{\tilde V_i\}_{i=1}^N)$ in the notation of Section~\ref{sec:conform_llm}
 
Recall that $\beta$ is the factuality threshold from~\eqref{eq:score_long}. We call a set of claims
$A \subseteq L(x)$ factually correct for $y$ if $w(c, y) \ge \beta$ for all $c \in A$.
 
\begin{theorem}
  Let $\{(X_i, Y_i)\}_{i=1}^{N+1}$ be exchangeable conditionally on $\mathcal{D}_{\mathrm{cal1}}$
  and assume that $\tilde V_1, \dots, \tilde V_{N+1}$
  are almost surely distinct. Then, for $\alpha \in \big[\frac{1}{N+1}, 1\big)$,
  \[
    1 - \alpha \le
    P\big(\forall c \in \bar L_\alpha(X_{N+1}),\ w(c, Y_{N+1}) \ge \beta\big)
    < 1 - \alpha + \frac{1}{N+1}.
  \]
\end{theorem}
 
In particular, the theorem applies to Algorithm~1, where $\hat\tau$ is the conditional quantile
regressor fitted on $\mathcal{D}_{\mathrm{cal1}}$ and $f_t(v) = t\,v$.

\begin{table}[t!]
\centering
\begin{tabular}{lccc}
\toprule
Ablation component & Adaptive wins & Mean bias reduction & Bias range \\
\midrule
Number of PCA components & 4/4 & 74.3\% & 0.52--1.49 \\
Embedding model & 4/4 & 80.9\% & 0.52--0.84 \\
$\mathcal{D}_{\mathrm{cal}_1}/\mathcal{D}_{\mathrm{cal}_2}$ split & 5/5 & 81.9\% & 0.52--0.74 \\
MLP configuration & 5/5 & 74.4\% & 0.42--1.90 \\
\bottomrule
\end{tabular}
\caption{Ablation on long-form QA (Mistral 7B, 50 random splits, target coverage 80\%): number of configurations where Adaptive has a lower coverage bias than Original, mean relative bias reduction, and bias range. Original: 3.63, oracle Group-wise: 0.36.}
\label{tab:ablation}
\end{table}

\begin{proof}
Conditionally on $\mathcal{D}_{\mathrm{cal1}}$, $\hat\tau$ is a fixed function, so the same map
$(x, y) \mapsto f^{-1}_{\hat\tau(x)}\big(V(x, y)\big)$ is applied to each pair $(X_k, Y_k)$.
Therefore, the transformed scores $\tilde V_1, \dots, \tilde V_{N+1}$ are exchangeable
conditionally on $\mathcal{D}_{\mathrm{cal1}}$. Below, all probabilities are conditional on
$\mathcal{D}_{\mathrm{cal1}}$ and since the final bounds do not depend on $\mathcal{D}_{\mathrm{cal1}}$,
they also hold unconditionally.
 
Let $k_\alpha := \lceil (N+1)(1-\alpha) \rceil$. Since $\frac{1}{N+1} \le \alpha < 1$, we have
$k_\alpha \in \{1, \dots, N\}$, so $N+1-k_\alpha \in \{1, \dots, N\}$, i.e., the threshold is one of
the calibration scores:
\[
  \hat q = \tilde V_{(N+1-k_\alpha)},
\]
where $\tilde V_{(1)} < \dots < \tilde V_{(N)}$ are the order statistics of $\tilde V_1, \dots, \tilde V_N$.
 
By~\eqref{eq:filtration}, the filtered set $F_t(x, L(x))$ is factually correct for $y$ if and only if $s(c) > t$ for every incorrect claim $c \in L(x)$, i.e., every claim with $w(c, y) < \beta$.
Hence, by~\eqref{eq:score_long}, $V(x, y)$ is the smallest uncertainty score among the incorrect claims,
and $F_t(x, L(x))$ is factually correct for $y$ if and only if $t < V(x, y)$.
Since $f_{\hat\tau(x)}$ is strictly increasing, $\bar L_\alpha(x) = F_t(x, L(x))$ with
$t = f_{\hat\tau(x)}(\hat q)$. Taking $(x, y) = (X_{N+1}, Y_{N+1})$ and applying
$f^{-1}_{\hat\tau(X_{N+1})}$ to both sides of $t < V(X_{N+1}, Y_{N+1})$, we obtain
\[
  \big\{\forall c \in \bar L_\alpha(X_{N+1}),\ w(c, Y_{N+1}) \ge \beta\big\}
  = \big\{\tilde V_{N+1} > \tilde V_{(N+1-k_\alpha)}\big\}.
\]
 
The event on the right means that $\tilde V_{N+1}$ is among the $k_\alpha$ largest of
$\tilde V_1, \dots, \tilde V_{N+1}$. Since these scores are exchangeable and almost surely
distinct, the rank of $\tilde V_{N+1}$ among them is uniformly distributed over
$\{1, \dots, N+1\}$. Therefore,
\[
  P\big(\tilde V_{N+1} > \tilde V_{(N+1-k_\alpha)}\big) = \frac{k_\alpha}{N+1}.
\]
 
Finally, by the definition of $k_\alpha$,
\[
  (N+1)(1-\alpha) \le k_\alpha < (N+1)(1-\alpha) + 1.
\]
Dividing by $N+1$ yields
\[
  1 - \alpha \le \frac{k_\alpha}{N+1} < 1 - \alpha + \frac{1}{N+1}.
\]
Combining the above proves the result.
\end{proof}
 

\section{Additional Experimental Results}
\label{app:additional_experiments}

\subsection{Ablation Study.}
\label{subsec:ablation}


We vary one component at a time (Mistral 7B, 50 random splits, $1-\alpha=0.8$): the number of principal components (4--32), the embedding model (QA-MPNet, MPNet, DistilRoBERTa, MiniLM), the $\mathcal{D}_{\mathrm{cal}_1}/\mathcal{D}_{\mathrm{cal}_2}$ split (25/45 to 45/25), and the MLP configuration.
Adaptive reduces the coverage bias of Original ($3.63$) in every configuration (Table~\ref{tab:ablation}) and is robust to the embedding model and the split; more components or a deeper network increase the variance of $\hat\tau$, while early stopping underfits it (bias $1.90$).

\subsection{Efficiency Metrics.}
\label{subsec:long_res}

\begin{table}[t]
\centering
\small
\setlength{\tabcolsep}{5pt}
\begin{tabular}{lcccc}
\toprule
& \multicolumn{2}{c}{Without category labels} & \multicolumn{2}{c}{With category labels} \\
\cmidrule(lr){2-3} \cmidrule(lr){4-5}
Category & Original & Adaptive & Group-wise & Boosted \\
\midrule
Inventions & 87.49 $\pm$ 0.89 & \good{83.19 $\pm$ 3.02} & 81.69 $\pm$ 1.60 & 83.68 $\pm$ 5.63 \\
Biographies & 82.46 $\pm$ 0.86 & \good{81.49 $\pm$ 2.35} & 81.39 $\pm$ 1.66 & 85.38 $\pm$ 5.88 \\
Artworks & \good{90.12 $\pm$ 0.64} & 92.05 $\pm$ 1.65 & 92.00 $\pm$ 0.94 & 91.81 $\pm$ 4.16 \\
Books & 85.17 $\pm$ 1.08 & \good{83.73 $\pm$ 2.49} & 82.21 $\pm$ 1.83 & 86.20 $\pm$ 5.84 \\
Cities & \good{82.50 $\pm$ 0.81} & 83.26 $\pm$ 2.29 & 83.00 $\pm$ 1.01 & 87.52 $\pm$ 4.76 \\
Movies & 83.90 $\pm$ 0.87 & \good{80.53 $\pm$ 3.24} & 82.21 $\pm$ 1.42 & 82.40 $\pm$ 8.04 \\
Landmarks & \good{82.01 $\pm$ 1.12} & 85.49 $\pm$ 2.32 & 86.58 $\pm$ 1.65 & 86.14 $\pm$ 6.76 \\
Events & \good{81.46 $\pm$ 1.18} & 83.84 $\pm$ 3.27 & 83.49 $\pm$ 1.88 & 83.15 $\pm$ 6.90 \\
\bottomrule
\end{tabular}
\caption{Fraction of removed claims (\%) for Mistral 7B on long-form QA at target coverage $80\%$ (mean $\pm$ std over seeds).
Color marks the label-free method that removes fewer claims.}
\label{tab:mistral_ccp_removed}
\end{table}

\begin{table}[t]
\centering
\small
\setlength{\tabcolsep}{5pt}
\begin{tabular}{lcccc}
\toprule
& \multicolumn{2}{c}{Without category labels} & \multicolumn{2}{c}{With category labels} \\
\cmidrule(lr){2-3} \cmidrule(lr){4-5}
Category & Original & Adaptive & Group-wise & Boosted \\
\midrule
Inventions & 88.06 $\pm$ 0.93 & \good{82.98 $\pm$ 2.58} & 82.24 $\pm$ 2.56 & 89.03 $\pm$ 3.80 \\
Biographies & \good{79.67 $\pm$ 0.94} & 82.63 $\pm$ 2.88 & 81.43 $\pm$ 1.25 & 90.74 $\pm$ 4.63 \\
Artworks & \good{81.14 $\pm$ 0.87} & 88.61 $\pm$ 1.97 & 87.92 $\pm$ 1.03 & 93.31 $\pm$ 3.73 \\
Books & 84.05 $\pm$ 0.87 & \good{79.85 $\pm$ 2.89} & 79.92 $\pm$ 1.52 & 87.11 $\pm$ 5.31 \\
Cities & 86.91 $\pm$ 0.63 & \good{84.11 $\pm$ 1.90} & 83.56 $\pm$ 1.08 & 89.40 $\pm$ 4.79 \\
Movies & 85.06 $\pm$ 0.79 & \good{80.59 $\pm$ 2.73} & 80.25 $\pm$ 1.43 & 87.80 $\pm$ 5.41 \\
Landmarks & 80.66 $\pm$ 0.97 & \good{79.48 $\pm$ 2.96} & 81.13 $\pm$ 1.68 & 85.06 $\pm$ 7.21 \\
Events & \good{78.67 $\pm$ 1.20} & 80.56 $\pm$ 3.27 & 81.63 $\pm$ 1.64 & 86.11 $\pm$ 4.76 \\
\bottomrule
\end{tabular}
\caption{Fraction of removed claims (\%) for Gemma-3 12B on long-form QA at target coverage $80\%$ (mean $\pm$ std over seeds).
Color: label-free method with fewer removals.}
\label{tab:gemma_removed}
\end{table}



Tables~\ref{tab:mistral_ccp_removed} and~\ref{tab:gemma_removed} report the fraction of removed claims for Mistral 7B and Gemma-3 12B.
As both methods are calibrated to the same marginal coverage, Adaptive removes about as many claims as Original overall and only redistributes the filtering across categories: it removes more claims where Original under-covers (e.g., Landmarks for Mistral 7B and Artworks for Gemma-3 12B) and fewer where it over-covers (e.g., Inventions).
This yields the improved category-wise coverage in Tables~\ref{tab:mistral_ccp_merged} and~\ref{tab:gemma_ccp_merged} without increasing the overall amount of filtering.
Per prompt, Adaptive retains on average 3.73 and 4.82 claims (vs. 3.66 and 4.64 for Original), and its fraction of removed correct claims stays within one point of Original (82.0\% vs. 82.2\% and 80.7\% vs. 81.3\%).

\newpage

\subsection{Additional Multiple-choice QA Results}

\begin{figure*}[h!]
  \centering
  \begin{minipage}[t]{0.43\textwidth}  
    \includegraphics[width=\textwidth]{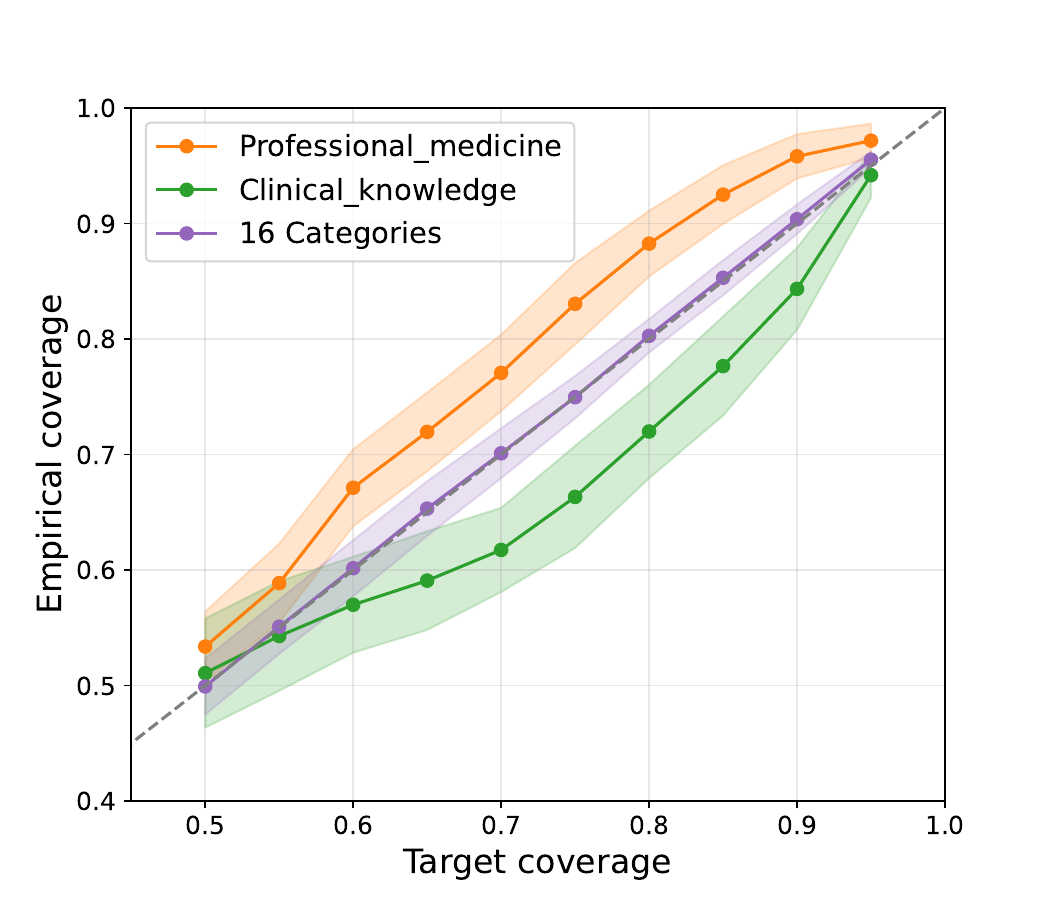}
    \vskip-10pt
    \centerline{(a)}
  \end{minipage}
  ~~~~
  \begin{minipage}[t]{0.43\textwidth}
    \includegraphics[width=\textwidth]{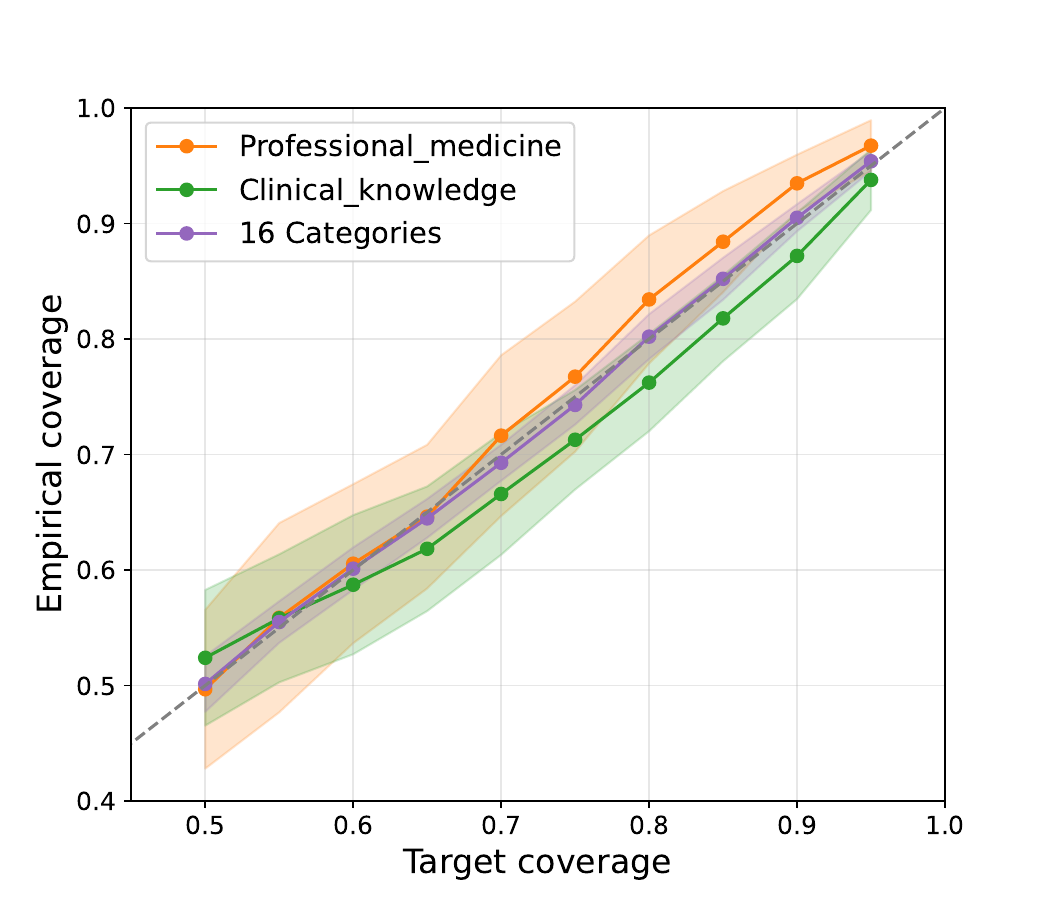}
    \vskip-10pt
    \centerline{(b)}
  \end{minipage}
  \vskip-5pt
  \caption{Multi-choice QA target vs. empirical coverage for (a) Conformal Factuality and (b) Our adaptive conformal method.
  The conformal procedure is calibrated on all 16 categories.
  While both methods achieve the target coverage marginally, only the adaptive approach closely approximates conditional coverage.
  }
  \vskip-10pt
\label{fig:mcqa_full}
\end{figure*}

This experiment uses Llama-2-7B with 20 random splits into three equal parts.
Figure~\ref{fig:mcqa_full} shows target versus empirical coverage for multi-choice QA when calibration is performed jointly across all $16$ categories.
While both methods achieve the desired marginal coverage overall, the global conformal approach exhibits substantial deviations at the category level, with over-coverage for Professional Medicine and under-coverage for Clinical Knowledge.
In contrast, the adaptive method produces curves that are closer to the diagonal for each category, indicating improved alignment with the target and better conditional coverage.

\subsection{Annotation Reliability.}

We validated GPT-4o factuality annotations through human evaluation of 40 claims sampled across all eight categories of long-form QA data.
GPT-4o achieved an F1 score of 0.898 against human judgments, with most disagreements arising from conservative labeling of correct claims as incorrect.
This suggests that annotation noise is limited.



\subsection{High-coverage Evaluation.}
Table~\ref{tab:high_coverage_longform} shows results for Mistral 7B for long-form QA at the high target coverage level of $1-\alpha=0.95$.
Even at this high-coverage level, the adaptive method remains effective.
The original conformal baseline frequently over-covers or under-covers individual categories, whereas adaptive method generally produces coverage values that are closer to the target level of 0.95.


\begin{table}[t]
\begin{tabular}{lccc}
\toprule
& \multicolumn{2}{c}{Without category labels} & With labels \\
\cmidrule(lr){2-3} \cmidrule(lr){4-4}
Category & Original & Adaptive & Group-wise \\
\midrule
Inventions & 97.84 $\pm$ 1.40 & \good{96.37 $\pm$ 2.48} & 95.76 $\pm$ 2.65 \\
Biographies    & 97.08 $\pm$ 1.50 & \good{94.23 $\pm$ 2.95} & 94.99 $\pm$ 2.80 \\
Artworks   & \good{94.89 $\pm$ 2.23} & 95.61 $\pm$ 2.92 & 95.43 $\pm$ 2.77 \\
Books      & 96.76 $\pm$ 1.71 & \good{95.35 $\pm$ 2.67} & 95.47 $\pm$ 2.58 \\
Cities     & 94.44 $\pm$ 2.08 & \good{95.27 $\pm$ 3.23} & 95.25 $\pm$ 2.51 \\
Movies     & 95.93 $\pm$ 1.86 & \good{94.28 $\pm$ 2.95} & 94.92 $\pm$ 2.68 \\
Landmarks  & 90.60 $\pm$ 2.77 & \good{94.67 $\pm$ 2.49} & 94.82 $\pm$ 2.99 \\
Events     & 92.69 $\pm$ 2.36 & \good{94.67 $\pm$ 2.75} & 95.64 $\pm$ 2.52 \\
\midrule
Cov. bias  & 1.87 & \good{0.59} & 0.35 \\
\bottomrule
\end{tabular}
\centering
\small
\caption{Coverage (\%) on long-form QA for Mistral~7B at target coverage $1-\alpha=0.95$. Results are reported as mean $\pm$ standard deviation over 100 random seeds. Color indicates the method whose empirical coverage is closer to the target value of 95\%.}
\label{tab:high_coverage_longform}
\end{table}


\section{Datasets}

\subsection{Long-form QA}
\label{app:dim_reduct}
\begin{figure*}[h]
  \centering
  \includegraphics[width=0.45\textwidth]{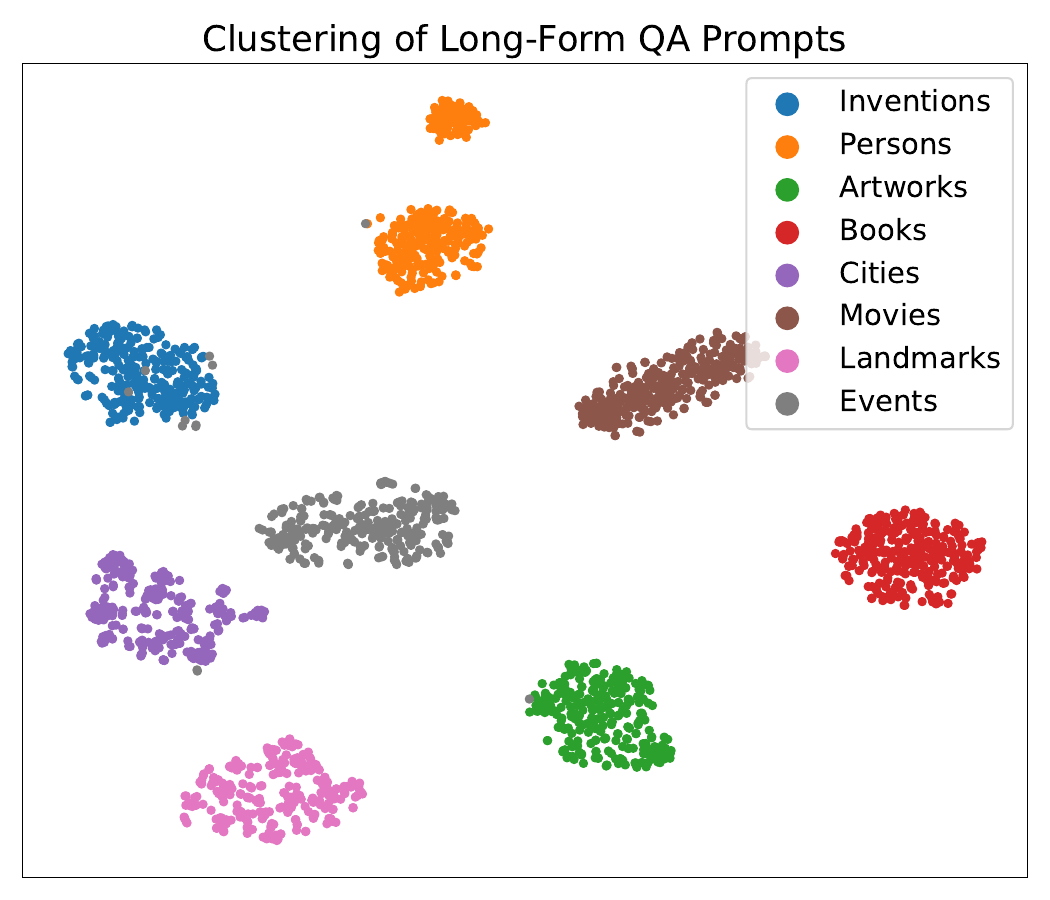}
  \caption{T-SNE visualization of PCA clustering of long-form QA prompts.}
  \vskip-10pt
\label{fig:long_clusters}
\end{figure*}

Figure~\ref{fig:long_clusters} shows a t-SNE visualization of PCA-reduced embeddings of long-form QA prompts, colored by category.
The prompts form well-separated clusters corresponding to different semantic categories, indicating that the embedding space captures meaningful differences between domains.

\end{document}